\documentclass[letterpaper, 10 pt, conference]{ieeeconf}  

\IEEEoverridecommandlockouts                              

\usepackage[utf8]{inputenc}
\usepackage[T1]{fontenc}
\usepackage{url}
\usepackage{listings}
\usepackage{graphicx}
\usepackage{multirow}
\usepackage[caption=false]{subfig}
\usepackage{booktabs}
\usepackage{mathtools}
\usepackage{tikz}
\usepackage{textcomp}
\usepackage{hyperref}
\usepackage{lipsum}

\usepackage{algorithm}
\usepackage[noend]{algpseudocode}
\usepackage{stmaryrd}

\let\labelindent\relax
\usepackage{enumitem}

\usepackage{calc}
\usepackage{amsmath,amssymb,amsfonts,amsthm}
\usepackage[noadjust]{cite}
\usepackage{textcomp}
\graphicspath{{Figures/}}

\newcommand\copyrighttext{%
  \footnotesize \textcopyright 2021 IEEE. Personal use of this material is permitted.
  Permission from IEEE must be obtained for all other uses, in any current or future
  media, including reprinting/republishing this material for advertising or promotional
  purposes, creating new collective works, for resale or redistribution to servers or
  lists, or reuse of any copyrighted component of this work in other works.
  DOI: t.b.d.}
\newcommand\copyrightnotice{%
\begin{tikzpicture}[remember picture,overlay]
\node[anchor=south,yshift=10pt] at (current page.south) {\fbox{\parbox{\dimexpr\textwidth-\fboxsep-\fboxrule\relax}{\copyrighttext}}};
\end{tikzpicture}%
}

\newenvironment{figfont}{\fontfamily{phv}\selectfont\footnotesize}{\par}
        
\begin{document}
\title{Safe, Deterministic Trajectory Planning for Unstructured and Partially Occluded Environments
\thanks{This project has received funding from the European Union’s Horizon 2020 research and innovation programme under grant agreement No 956123 - FOCETA.}%
\thanks{$^{1}$S. vom Dorff, M. Kneissl are with the Corporate R\&D department of DENSO Automotive Deutschland GmbH, Freisinger Str. 21-23, 85386 Eching, Germany
        {\tt\small \{s.vomdorff, m.kneissl\}@eu.denso.com}}%
\thanks{$^{2}$M. Fränzle and S. vom Dorff are with the Carl von Ossietzky University, Department of Computing Science, 26111 Oldenburg, Germany
        {\tt\small fraenzle@informatik.uni-oldenburg.de}}%
}
\author{Sebastian vom Dorff$^{1,2}$, Maximilian Kneissl$^{1}$, Martin Fränzle$^{2}$}
\date{April 2021}

\maketitle

\copyrightnotice

\begin{abstract}
Ensuring safe behavior for automated vehicles in unregulated traffic areas poses a complex challenge for the industry.
It is an open problem to provide scalable and certifiable solutions to this challenge. 
We derive a trajectory planner based on model predictive control which interoperates with a monitoring system for pedestrian safety based on cellular automata.
The combined planner-monitor system is demonstrated on the example of a narrow indoor parking environment.
The system features deterministic behavior, mitigating the immanent risk of black boxes and offering full certifiability.
By using fundamental and conservative prediction models of pedestrians the monitor is able to determine a safe drivable area in the partially occluded and unstructured parking environment.
The information is fed to the trajectory planner which ensures the vehicle remains in the safe drivable area at any time through constrained optimization.
We show how the approach enables solving plenty of situations in tight parking garage scenarios.
Even though conservative prediction models are applied, evaluations indicate a performant system for the tested low-speed navigation.
\end{abstract}
\section{Introduction}
The challenge of creating a symbiosis between safety and performance in automated vehicles is a topic of passionate discussions and research.
While the promises of an accident-free world invite to dream about the future, it is hard to even prove that automated vehicles can be on par with human drivers \cite{nidhi_kalra_susan_m_paddock_driving_nodate}.
The expectations vary from being as good as human drivers up to a thousand times better and with it the complexity of models to identify risky situations before they turn hazardous \cite{shalev-shwartz_formal_2017}.

Especially the interaction in less clearly defined environments poses a challenge to algorithms as they cannot simply guess the intentions of their environment. This problem grows bigger when dealing with vulnerable road users which can quickly change their speed in directions, most prominently pedestrians \cite{koschi_set-based_2018}. Consequently, numerous modeling approaches have emerged with varying level of detail and complexity \cite{lefevre_survey_2014, camara_pedestrian_2020}.
The first thought would obviously be to pick the most sophisticated model available but that bears two shortcomings:
\begin{itemize}
\item The certifiability is not equally given for all models. Especially machine learning approaches lack the possibility to be formally verified, thus requiring enormous test driving for validation \cite{oh_towards_nodate, wachenfeld_how_nodate}. This is not only an economic problem due to being obliged to drive these miles over and over again for any change in the algorithms. To make matters worse, they would ultimately lead to fatal accidents just for the sake of proving their statistical safety.
\item Available computation power is a technical problem in the way to automated vehicles. Especially with the rise of electric vehicles, the limitation of energy proves a substantial hurdle for the adaption of power intensive algorithms. Furthermore, the pure hardware costs have a significant influence on the real world adaption.
\end{itemize}

Experiences from other safety related engineering tasks have shown that a division of responsibilities ease the requirements on each module. This allows to exploit the strengths of different approaches while mitigating the corresponding shortcomings.
Within the most typical implementations for that sake are actor-monitor configurations as being commonly known from industries such as aerospace \cite{kornecki_approaches_2004}. While those approaches usually tend to be fail-safe, i.e. making sure an error can be solved by a human supervisor without the machine interfering in a hazardous way, this option is not available in fully automated vehicles \cite{koopman_challenges_2016}.

For this reason, the automation must consider the case of failure before acting. Just like human drivers are asked to adjust their speed so that they can come to a full stop within the visible area, the automated vehicle must intrinsically provide an emergency plan in case of system failures \cite{pek_online_2019}.

This can be achieved by optimization-based control schemes which can ensure that a vehicle's trajectory remains in a safe space at any time through the consideration of constraints.
The predictive nature of such methods additionally allows to plan to a safe stop at the end of each prediction interval.
Overall, this strategy suggests an important contribution to the system's safety argumentation \cite{vom2020fail}.

Additionally, the automated vehicle needs to anticipate hazardous situations which can be caused by other traffic participants which cannot be directly perceived as they might be occluded by other objects \cite{orzechowski_tackling_2018}. This problem enhances in the case of unstructured environment such as parking garages.

In our approach we are focusing on actor-monitor configurations with deterministic behavior. To avoid overly demanding computation effort, the model features a pessimistic physical pedestrian model which calculates the free-space as being used by the model predictive control as planning constraints.
This is based on a combination of ray-tracing algorithms to create a field of view (FoV) and a cellular automaton extrapolating pedestrian positions also from occluded areas, enabling decision making under uncertainty due to limited vision. 
The monitor is acting independent of the controller loop and can only accept or reject a given trajectory by the nonlinear model predictive controller (NMPC).
We show that even a pessimistic and simplistic environment model, in this case focusing on pedestrians, ensures safe behavior while maintaining acceptable performance for most driving scenarios in parking environments. We therefore provide the benefit of only being forced to apply more complex approaches in rare cases when the performance provided is not satisfying.
Furthermore, we evaluate how far the capabilities of such a system reach before being forced to degrade performance drastically.

The remainder of the paper is organized as follows.
In the next section the general set-up of our approach is schematically presented and explained. Followed by that in Section III, a number of experiments are shown to assess the performance. The results of the experiments are discussed in Section IV. Lastly a final conclusion is given in Section V.
\section{Approach}

\begin{figure}
\def\svgwidth{\columnwidth}
\begin{center}
\begin{figfont}
\begin{tiny}
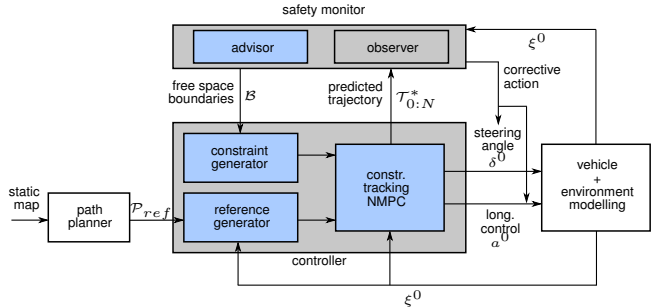
\end{tiny}
\end{figfont}
\end{center}
\vspace{-5mm}
\caption{Safety and control architecture.}
\label{fig:architecture}
\end{figure}

\subsection{Safety Monitor}
The safety monitor as seen in the top box in Fig. \ref{fig:architecture} is divided into two parts:

\subsubsection{Advisor}
\begin{figure}
\begin{center}
\includegraphics[width=0.4\textwidth]{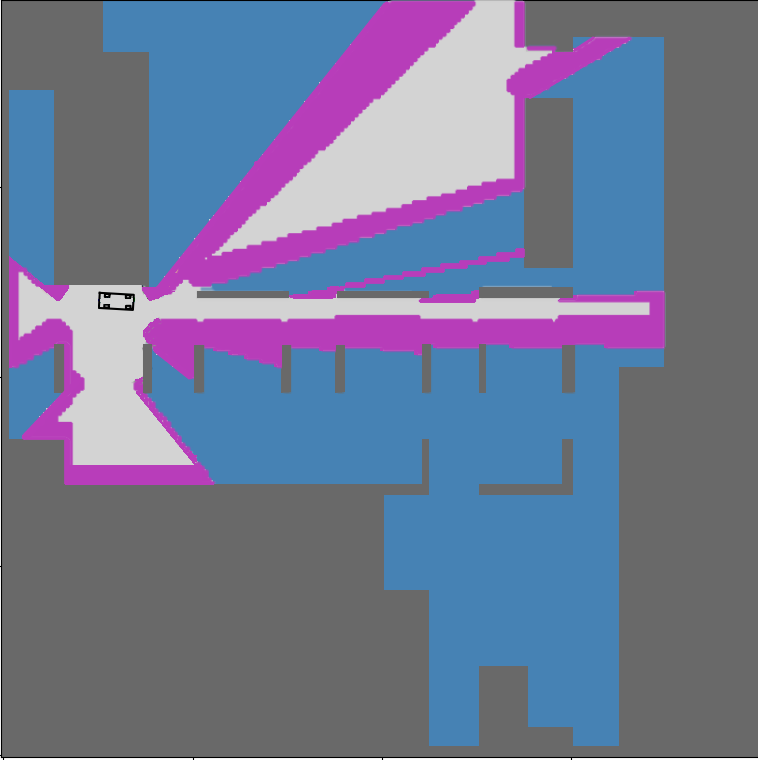}
\end{center}
\caption{The free space estimation process by extrapolation. The static garage walls are depicted in dark gray, the hidden areas in blue and the extrapolated movement areas in purple. Light gray areas are drivable safely.}
\label{fig:fov_combined}
\end{figure}

\begin{figure}
\def\svgwidth{0.4\columnwidth}
\begin{center}
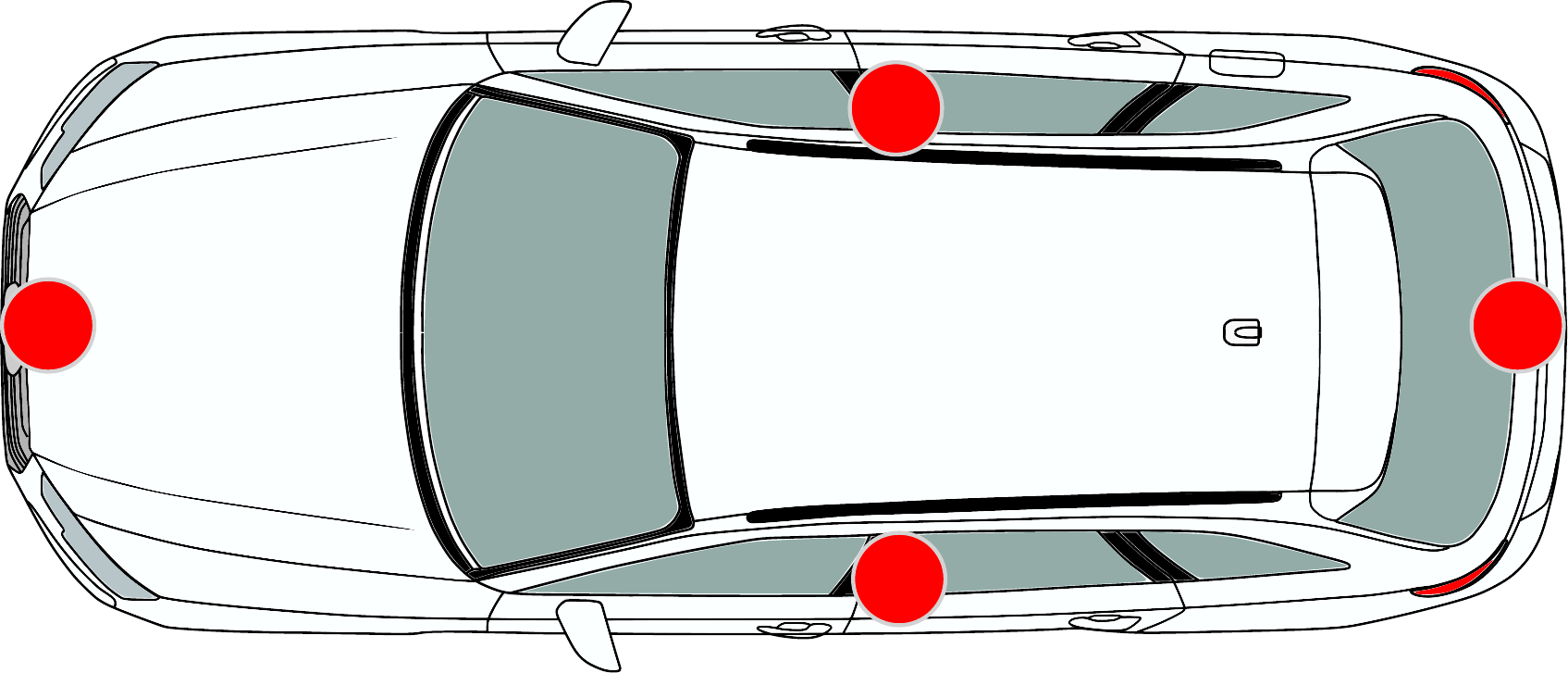
\end{center}
\caption{Sensor positions on the vehicle.}
\label{fig:sensor_position}
\end{figure}

The advisor is meant to provide the NMPC with enough information to generate a trajectory, without sharing the whole intrinsic data to prevent a specific adoption by the planner. This would jeopardize the independence of the two modules.
Internally, the advisor is a free-space detector that operates on an occupancy grid map (OGM) depicting the parking environment with its static properties such as drivable areas, walls and potential entrances for pedestrians.
The functionality is designed as follows:
\begin{enumerate}
\item On a given map, as seen by the gray structures in Fig. \ref{fig:fov_combined}, a surround vision around the vehicle is reconstructed. Four sensors which are assumed to have perfect perception within the FoV are located at the outer points of the vehicle i.e., at the front center, rear center, and on the left and right b-pillar as shown in Fig.\ref{fig:sensor_position}.
\item With those virtual sensors a FoV map $\mathcal{F}$ is generated, indicating which areas of the map are hidden. This is done by utilizing the Bresenham algorithm \cite{bresenham1965algorithm} to ray cast from the position of the sensor to the first collision point of the ray with the static environment.
As an outcome the OGM cells which are found to be outside of the FoV are marked as being occupied by a pedestrian as well as those cells which have actually been detected to inhabit a pedestrian. Those are the blue areas of Fig. \ref{fig:fov_combined}.
\item With the FoV and the pedestrians marked in the map, a customized cellular automaton with variable update-rate is used to extrapolate the potential positions of the pedestrians for a certain time interval. This is shown in purple in Fig. \ref{fig:fov_combined}. The time frame of the extrapolation is requested by the NMPC and can be freely adjusted. The extrapolation is done by assuming a maximal movement speed for pedestrians which is limited by
\begin{equation}
v_{ped} \leq \min{(\overline{v}_{ped}, v_0)},
\end{equation} 
with $v_0$ being the current vehicle speed and $\overline{v}_{ped}$ being a maximum expected speed for pedestrians.
This requirement is introduced as it can be considered unreasonable if the pedestrian hits the vehicle with a faster speed than the vehicle is driving itself. An exception from this are pedestrians which have already been identified and tracked with a certain speed. In that case, the actual observed speed of the pedestrian is used.
When the extrapolation has been completed, the OGM features the safe drivable area $\mathcal{M}$, depicted as light gray in Fig. \ref{fig:fov_combined}.
\end{enumerate}

With the help of this generated free space map, the free distance in different directions of the vehicle is detected. In the given work these are the straight lines from the vehicle to the front, rear, left, and right. Anyhow, the module allows to determine the free distance for any amount of directions, such as one sample for every degree. This is done using the same ray casting algorithm as mentioned before but considering the dynamic map and only being able to search in certain directions.
Additionally, the NMPC can request to leave a certain buffer to the potentially occupied cells, such as one or two meters of margin.
Eventually, the free space detector returns the final free coordinate pairs $\mathcal{B}$ in the requested directions to the controller.

\subsubsection{Observer}
The observer is a collision checking module. It receives the extrapolated map with potential pedestrian positions and the planned path of the NMPC which is required to reach a full stop at the last position of the trajectory.
It then superimposes the OGM with the trajectory and judges if potentially occluded cells overlap with the trajectory. In that case the trajectory is rejected, and the vehicle controller continues the execution of its latest received accepted trajectory that will potentially end with a full stop. 
If there is no collision between the extrapolated map and the trajectory, the trajectory is accepted for the actual vehicle control.
Due to the nature of the underlying processes, the combination of these two modules can be run on lightweight, dedicated hardware.

\subsection{Controller}
The controller's task is to compute a safe and comfortable trajectory candidate which predicts the future vehicle movement and provides longitudinal and lateral control signals.
The trajectory candidate will be checked against a safety specification as described in Section II-A-2.
This part is out of scope for this paper.
It is assumed that the solution of the nominal constraint NMPC fulfills the safety specification by providing trajectories within the environment's free space that is shared by the safety advisor.
Suitable observer strategies have been discussed in \cite{vom2020fail}.
Additional to the free space boundaries $\mathcal{B}$, the controller receives a reference path $\mathcal{P}_{ref}$ in form of waypoints and its initial condition as inputs.
In the following, we describe the setup of the optimization problem that is used for the NMPC controller.
Fig. \ref{fig:architecture} visualizes the controller architecture.

\subsubsection{Vehicle Model}
The vehicle is modelled using the following discretized nonlinear kinematic bicycle model, derived from \cite{kong2015kinematic} and illustrated in Fig. \ref{fig:model}:
\begin{figure}
\def\svgwidth{0.45\textwidth}
\begin{center}
\includegraphics[height=4cm]{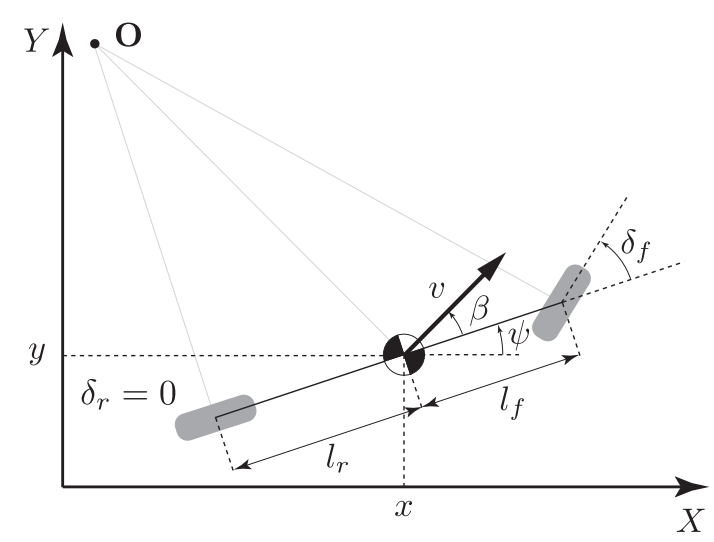}
\end{center}
\caption{Nonlinear kinematic bicycle model \cite{kong2015kinematic}.}
\label{fig:model}
\end{figure} 
\begin{equation}
\label{eq:model}
\xi^{t+1} = 
\begin{pmatrix}
x^{t+1} \\
y^{t+1} \\
\psi^{t+1} \\
v^{t+1}
\end{pmatrix}
=
\begin{pmatrix}
x^{t} \\
y^{t} \\
\psi^{t} \\
v^{t}
\end{pmatrix}
+
\begin{pmatrix}
v^{t} \cos(\psi^{t} + \beta^{t}) \\
v^{t} \sin(\psi^{t} + \beta^{t}) \\
\frac{v^{t}}{l_r} \sin(\beta^{t}) \\
a^{t}
\end{pmatrix}T_s,
\end{equation}
with $\beta = \tan^{-1}\left(\frac{l_r}{l_f + l_r} \tan(\delta^{t})\right)$.
In \eqref{eq:model}, $t$ is a discrete time step and $T_s$ the sampling time, in the state vector $\xi^t \in \mathbb{R}^4$ the sates $x^{t}$ and $y^{t}$ are the center of gravity (CoG) coordinates in a global $X-Y-$frame, $\psi^{t}$ is the vehicle's heading in the same frame, $v^{t}$ the CoG velocity, $\beta^{t}$ its angle, $a^{t}$ its acceleration, $l_f$ and $l_r$ are the distance between the CoG and the front and rear axle, respectively, and $\delta^{t}$ is the steering angle. 
The control inputs are $a^t$ and $\delta^t$.
It has been shown in \cite{polack2018guaranteeing} that this model delivers accurate predictions for low-speed maneuvers.

\subsubsection{Reference Generation}
The reference path is provided to the controller as a set of $n$ waypoints in the global coordinate frame,
\begin{equation}
\label{eq:path}
\begin{split}
\mathcal{P}_{ref} = \lbrace & \left( x_{p,1}, y_{p,1}, h_{p,1}, \kappa_{p,1} \right)^T, \\
 & ..., \left( x_{p,n}, y_{p,n}, h_{p,n}, \kappa_{p,n} \right)^T \rbrace,
\end{split}
\end{equation}
defined by the $X-Y$ position $x_{p,\cdot}$ and $y_{p,\cdot}$, as well as the respective reference heading $h_{p,\cdot}$, and a path reference curvature $\kappa_{p,\cdot}$.
The Euclidean distance between consecutive waypoints is assumed to be constant, i.e.,
\begin{equation}
\label{eq:dist_wp}
\begin{split}
\Delta p = \sqrt{(x_{p,i} - x_{p,j})^2 + (y_{p,i} - y_{p,j})^2} = const, \\
\ i \in [1..n-1], j = i+1,
\end{split}
\end{equation}
where $[1..n-1] = \{1,2,...,n-1\}$. 
We define a reference velocity
\begin{equation}
\label{eq:v_ref_path}
v_{p,i} = \sqrt{\frac{\overline{a}_{lat}}{|\kappa_{p,i}|}},\ i \in [1..n],
\end{equation}
for each waypoint with the design parameter $\overline{a}_{lat}>0$ representing the maximum allowed lateral vehicle acceleration.
The initial reference state for a time step $t = 0$ is defined as
\begin{equation}
	\label{eq:ref_init}
	\xi_{ref}^{t} = \left( x_{p,i}, y_{p,i}, h_{p,i}, v_{p,i} \right)^T,
\end{equation}
with 
\begin{equation}
	\label{eq:initial_index}
	i = \underset{j \in [1..n]}{\text{argmin}} \sqrt{(x_{p,j} - x^0)^2 + (y_{p,j} - y^0)^2}.
\end{equation}
Given \eqref{eq:path} - \eqref{eq:initial_index} and let $i$ be the path index for time step $t-1$.
Then, a consecutive reference vector according \eqref{eq:ref_init} with an index $j$ is defined for time steps $t>0$ with
\begin{equation}
	\label{eq:consecutive_index}
	j = i + \bigg\lfloor \frac{v_{p,i}T_s}{\Delta p} \bigg\rfloor,
\end{equation}
where $T_s$ is the controller's sampling time.

\subsubsection{Constraint Generation}
\begin{figure}
\vspace{3mm}
\def\svgwidth{0.45\textwidth}
\begin{center}
\begin{figfont}
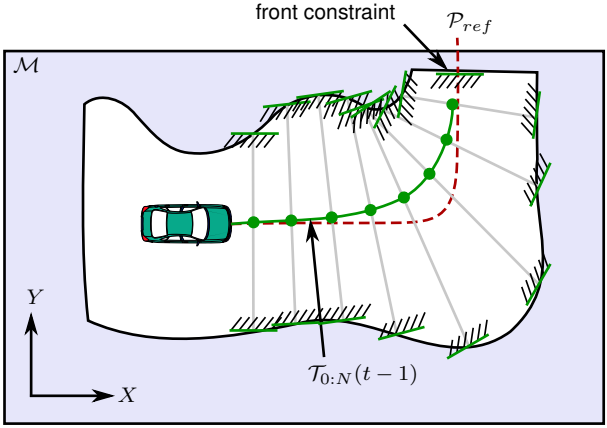
\end{figfont}
\end{center}
\vspace{-5mm}
\caption{Generation of environment constraints.}
\label{fig:constraints}
\end{figure}

The foremost target of the constraint formulation in this subsection is to provide the optimization problem of the NMPC with information about the safe space that the vehicle is allowed to move in.
First, a trajectory at time step $t$ is defined as
\begin{equation}
	\label{eq:trajectory}
	\mathcal{T}_{0:N}(t) = \bigg( \Big(\big(\xi^0\big)^T, \delta^0, a^0\Big), ..., \Big(\big(\xi^N\big)^T, \delta^N, a^N\Big) \bigg),
\end{equation}
with a horizon length $N$ and $\delta^N = a^N = 0$.
The safety advisor provides the boundary points $\mathcal{B}$ that defines the drivable space.
Based on this information a set of constraints to limit the vehicles positions is determined.
The procedure is illustrated in Fig. \ref{fig:constraints}.
Given the trajectory computed at the previous time step, $\mathcal{T}_{0:N}(t\!-\!1)$, we determine two half-space constraints for each stage of the trajectory.
Inspired by \cite{liniger2015optimization}, we achieve a set of constraints of the form
\begin{equation}
	\label{eq:half_space_constraints}
	P^t
	\begin{pmatrix}
		x^{t} \\
		y^{t}
	\end{pmatrix}
	\leq
	\gamma^t, \ t \in [1..N],
\end{equation}
with ${P^t \in \mathbb{R}^{2 \times 2}}$ and ${\gamma^t \in \mathbb{R}^2}$.
The two rows in \eqref{eq:half_space_constraints} determine the left and right bounds for each trajectory stage, respectively, and form a convex set.
The heading of the half-spaces is aligned with the respective trajectory stage heading $\Psi^t$.
Besides left and right constraints, a front constraint (cp. Fig.~\ref{fig:constraints})  is formulated such that the predicted trajectory does not leave the drivable area in forward direction.
This is achieved with a similar format as \eqref{eq:half_space_constraints} but with a perpendicular heading with respect to the reference path and only a single half-space constraint.

Additionally, we define box constraints on the remaining vehicle states and inputs as follows:
\begin{align}
	|\psi^{t+1} - \psi^{t}| & \leq \Delta \psi_{max} \label{eq:box_constraints1} \\
	v_{min} \leq v^t & \leq v_{max} \\
	\delta_{min} \leq \delta^t & \leq \delta_{max} \\
	a_{min} \leq a^t & \leq a_{max} \label{eq:box_constraints4},
\end{align}
where $\Delta \psi_{max}$ is a limit on the heading rate and $v_{min}$ and $v_{max}$ are the lower and upper bound for the velocity state, respectively, and similar for the control inputs. 

\subsubsection{NMPC Formulation}
\label{sec:NMPC_formulation}
The NMPC solves a finite horizon nonlinear optimization problem in discrete time at each sampling time step.
It applies the resulting control inputs from the solution of the first horizon step, and repeats this procedure with an updated initial state for consecutive time steps. Thus, a receding horizon methodology is achieved.

The nonlinear optimization problem is formulated as a constraint reference tracking problem:
\begin{align}
	\label{eq:MPCproblem}
	\mathcal{T}_{0:N}^{*}(t) & = \underset{\mathcal{T}_{0:N}(t)}{\text{argmin}}
	\sum_{t=0}^{N} \big\Vert \xi^t - \xi_{ref}^t \big\Vert_{Q}^2 + \sum_{t=0}^{N-1} \big\Vert  (\delta^t,a^t)^T \big\Vert_{R}^2 \\
	& \text{s.t.}\ \ \eqref{eq:model}, \eqref{eq:box_constraints1} - \eqref{eq:box_constraints4}, \hspace{15mm} t \in [0..N\!-\!1] \nonumber \\
	& \hspace{7mm} \eqref{eq:half_space_constraints} \nonumber \\
	& \hspace{7mm} v^N = a^N = 0 \label{eq:MPCproblem_terminal},
\end{align}
with the weighting matrices $Q \in \mathbb{R}^{4 \times 4}$ and $R \in \mathbb{R}^{2 \times 2}$ being positive semi-definite and positive definite, respectively, and notation $\Vert \cdot \Vert_Q^2$ represents the weighted 2-norm.
\eqref{eq:MPCproblem_terminal} is a full stop condition that is required for the present safety architecture \cite{vom2020fail}.
The nonlinear optimization problem \eqref{eq:MPCproblem} is then solved with a sequential quadratic programming (SQP) strategy, while in each iteration sequence a quadratic program (QP) is solved \cite{nocedal2006sequential}.
The latter is achieved through a linearization of the nonlinear model \eqref{eq:model} around the trajectory stages of $\mathcal{T}^{*}_{0:N}(t-1)$ from the previous time step.
The system's overall algorithmic implementation is summarized in the following subsection.

\subsection{Algorithmic Implementation}
The combined safety advise and control implementation is presented in Algorithm~\ref{alg:SafetyControl}.

\begin{algorithm}
\caption{Safety-ensured Trajectory Computation}\label{alg:SafetyControl}
\begin{algorithmic}[1]
\State \textit{\textbf{Safety Monitor:}}
\State Receive FoV map $\mathcal{F}$ 
\State generate safe drivable map $\mathcal{M}$ by extrapolation with the cellular automaton
\State output free-space boundaries $\mathcal{B}$
\State \textit{\textbf{Controller:}}
\State Given $\mathcal{P}_{ref}$, determine the reference $\xi_{ref}^t$ with \eqref{eq:dist_wp} -- \eqref{eq:consecutive_index}.
\State Given $\mathcal{M}$, determine the constraint set \eqref{eq:half_space_constraints}
\State Given $\xi^0$, solve the NMPC \eqref{eq:MPCproblem}, share $\mathcal{T}^{*}_{0:N}$ with the safety monitor, apply $\delta^0$ and $a^0$ to the vehicle
\State repeat from Step 1
\end{algorithmic}
\end{algorithm}

\section{Experiments}
In order to assess the performance of the drawn approach, three example scenarios have been chosen:
\begin{enumerate}[label=(\Alph*)]
\item The vehicle enters an empty garage and navigates the way to the main parking area
\item The vehicle navigates through a garage that has parked vehicles and several pedestrians in it
\item The vehicle emerges from a parking lot where it is surrounded by other cars limiting its vision.
\end{enumerate}
The environment for these scenarios is a 2D OGM that has been derived from a real parking garage. The OGM consists of 200 times 200 cells with 0.5~m edge length each, thus depicting a real-world area of 10000~$m^2$.
The parameters for the NMPC are shown in Table \ref{tab:exp_parameters}.

\begin{table}[]
\caption{The model parameters used for the NMPC in the experiments.}
\label{tab:exp_parameters}
\begin{center}
\begin{tabular}{@{}ll@{}}
\toprule
Parameter                    & Value                       \\ \midrule
$T_s$                        & $0.1~s$                     \\
$l_r$                        & $1~m$                       \\
$l_f$                        & $1.82~m$                    \\
$\overline{a}_{lat}$         & $1.3~m/s^2$               \\
$N$                          & $15$                        \\
$\psi_{min}, \psi_{max}$     & $-\pi/4~rad, \pi/4~rad$           \\
$a_{min}, a_{max}$         & $-7~m/s^2, 3~m/s^2$     \\
$v_{min}, v_{max}$         & $0~m/s, 5.56~m/s$               \\
$\delta_{min}, \delta_{max}$ & $-\pi/4~rad, \pi/4~rad$            \\
$Q$                          & diag $(0.1, 0.1, 1.0, 0.5)$ \\
$R$                          & diag $(0.01, 0.1)$          \\ \bottomrule
\end{tabular}%
\end{center}
\end{table}

In the given experiments, the model constraints only considered the COG and ensured this single point to be collision free. Even though possible, no inflating of the constraint to the whole vehicle footprint was done. This enabled to judge the severity of the situation when the NMPC was not able to deliver a collision free solution for the COG anymore. To evaluate the results, we have chosen four categories of performance.
\begin{itemize}
\item \emph{safe}: The vehicle moves through the parking garage to its destination without getting in any conflict with potential pedestrians.
\item \emph{minor conflict}: The vehicle touches the potentially conflicting areas slightly but never with more than 10\% of the vehicle footprint. An accident would be easily avoided if the pedestrian just takes minimal evasive action i.e., reduces the speed towards the vehicle.
\item \emph{critical}: The vehicle enters areas which could lead to conflicts with potential pedestrians. There is a significant overlap of the vehicle with the conflicting area but not more than 50\% of vehicles footprint.
\item \emph{hazardous}: The vehicle enters with over 50\% of its footprint the conflicting area, posing a serious hazard to pedestrians. 
\end{itemize}

The implementation of the experiments was realized in Python 3.6 on a standard Ubuntu 18.04 machine with an Intel Core i7-6700K CPU @ 4.00GHz × 8 and 32GB RAM.
\subsection{Case A}
\begin{figure}
\begin{center}
\includegraphics[width=0.45\textwidth]{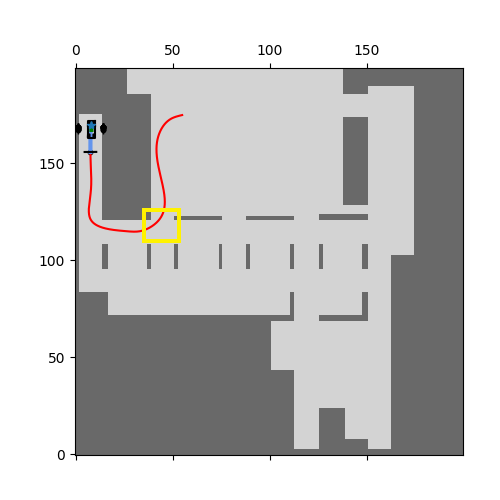}
\end{center}
\caption{The empty garage layout with the planned path and boundary points. The black box on the top left represents the vehicle's footprint. The red line depicts the overall planned path, while the blue line shows the intermediate trajectory. The black dots represent the boundaries given to NMPC. The yellow area points out the most challenging situation in the experiment.}
\label{fig:empty_garage}
\end{figure}

The first experiment covers the basic case of the vehicle driving from the entrance ramp towards the main parking area. There are no pedestrians in the parking area but the system is not aware of that. The experiment shows the ability of our approach to deal with narrow driveways and many occluded areas.
Fig. \ref{fig:empty_garage} shows the map layout and the planned path through it. The red path shows the overall track to the destination, while the blue trajectory is a slice of this path with an added speed-profile.

Consequently, the safety advisor extrapolates all potential areas that could be reached by pedestrians in a certain time frame from those occluded areas. From there, the boundary points are generated and fed back to the controller. Fig. \ref{fig:fov_combined} shows the final extrapolation of the map with the assumed parameters of $1.5~s$ time-horizon and a maximum pedestrian speed of $1~m/s$. The free map space is marked in light grey.

\begin{table}[]
\caption{The results from experiment A with different maximal pedestrian speeds $v=\overline{v}_{ped}$ and $T=N*T_s$.}
\label{tab:exp_A_eval}
\resizebox{\columnwidth}{!}{%
\begin{tabular}{@{}lllll@{}}
\toprule
          & $v=0.5~m/s$    & $v=1.0~m/s$    & $v=1.5~m/s$    & $v=2.0~m/s$    \\ \midrule
$T=1.0~s$ & safe           & safe           & safe           & minor conflict \\
$T=1.5~s$ & safe           & safe           & minor conflict & critical       \\
$T=2.0~s$ & safe           & minor conflict & critical       & hazardous      \\
$T=2.5~s$ & minor conflict & critical       & hazardous      & hazardous      \\ \bottomrule
\end{tabular}%
}
\end{table}

Table \ref{tab:exp_A_eval} shows the capability of the set-up to navigate without any incidents through the parking garage for experiment A. Several cases just require minor caution by the pedestrians to be solved. Besides, slight adjustments to the control or the extrapolation algorithm could already solve the scenarios in question. Most issues occurred at the area marked in yellow in Fig. \ref{fig:empty_garage}. The tight corner in combination with occluded spaces makes the handling of the situation difficult.

\subsection{Case B}
\begin{figure}
\begin{center}
\includegraphics[width=0.5\textwidth]{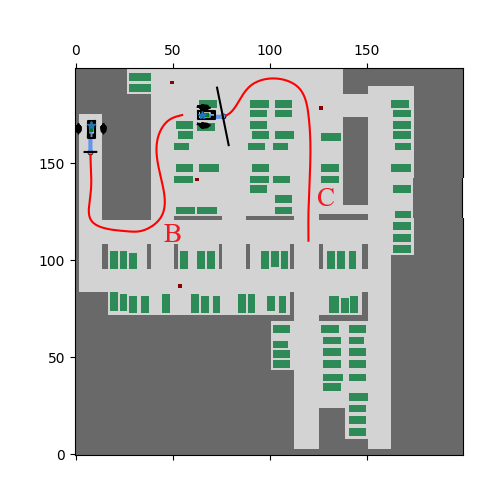}
\end{center}
\caption{The modified layout of the garage map. Added vehicles are marked in green, pedestrian position are shown in dark red. The bright red lines depict the vehicle's path in experiments B and C.}
\label{fig:experiment_b_and_c_layout}
\end{figure}

The second experiment, depicted by the red path "B" in Fig. \ref{fig:experiment_b_and_c_layout}, increases the difficulty of the first scenario by adding parked cars - for which we consider being large enough to hide pedestrians behind it - and several pedestrians which have been detected. In order to assess the capabilities of acting under uncertainty, the pedestrians do not feature any particular speed information, only their existence and position is determined beforehand.
Considering potential deviation in the detected position, pedestrians are placed with a $1~m$ times $1~m$ square surrounding them.

Table \ref{tab:exp2_eval} shows the results from experiment B. During the observation it could be noticed that the most critical spot on the path is still the second corner as it combines low visibility with plenty open spots that could potentially inhabit pedestrians.
The effect of other parking cars was only affecting the result when the time-horizon was selected at least with $2~s$. Otherwise the problematic situations still occurred in the before mentioned yellow area.

\begin{table}[]
\caption{The results from experiment B with different maximal pedestrian speeds $v=\overline{v}_{ped}$ and $T=N*T_s$.}
\label{tab:exp2_eval}
\resizebox{\columnwidth}{!}{%
\begin{tabular}{@{}lllll@{}}
\toprule
          & $v=0.5~m/s$    & $v=1.0~m/s$    & $v=1.5~m/s$    & $v=2.0~m/s$    \\ \midrule
$T=1.0~s$ & safe           & safe           & safe           & minor conflict \\
$T=1.5~s$ & safe           & safe           & minor conflict & critical       \\
$T=2.0~s$ & safe           & minor conflict & critical       & hazardous      \\
$T=2.5~s$ & minor conflict & critical       & hazardous      & hazardous      \\ \bottomrule
\end{tabular}%
}
\end{table}

\subsection{Case C}
As third case for the experiments we refer to the intuitively challenging task of leaving the parking lot while being surrounded by other vehicles limiting vision around the ego vehicle. In combination with the environment from case B it is the most challenging scenario to tackle. Furthermore, the path runs closer to pedestrians, evaluating how an open encounter with vulnerable traffic participants is handled.
The experiment layout is shown as path "C" in Fig. \ref{fig:experiment_b_and_c_layout}.


The results from this case are clearer than in the other two cases as Table \ref{tab:exp_C_eval} shows. There is only one critical case while the others are either safe or hazardous.
The most critical situation appeared shortly after leaving the parking lot since a blind spot behind the car parked next to the ego vehicle formed. At the same time, the vehicle speed increased, allowing the pedestrians to accelerate as well by definition.

\begin{table}[]
\caption{The results from experiment C with different maximal pedestrian speeds $v=\overline{v}_{ped}$ and $T=N*T_s$.}
\label{tab:exp_C_eval}
\resizebox{\columnwidth}{!}{%
\begin{tabular}{@{}lllll@{}}
\toprule
          & $v=0.5~m/s$    & $v=1.0~m/s$    & $v=1.5~m/s$    & $v=2.0~m/s$    \\ \midrule
$T=1.0~s$ & safe           & safe           & safe           & safe \\
$T=1.5~s$ & safe           & safe           & safe & hazardous       \\
$T=2.0~s$ & safe           & safe & hazardous       & hazardous      \\
$T=2.5~s$ & safe & critical       & hazardous      & hazardous      \\ \bottomrule
\end{tabular}%
}
\end{table}
\section{Discussion}
The experiments have shown the general capability of a conservative prediction model for enabling safe driving in parking environments. In a noteworthy range of parameter combinations, the approach delivered sufficient performance to fulfill different driving maneuvers within the parking environment.
With a planning time-horizon of $1~s$ to $1.5~s$, a fluent operation of a control algorithm with an emergency stop procedure is possible while maintaining safety under the premise that pedestrians do not exceed a certain speed. 
However, there are shortcomings where the system cannot guarantee smooth operation without sacrificing safety.
While the system is designed to never enter a conflicting zone, the model only considers a point particle for the simulation, leaving room for the vehicle footprint to interact with those zones. The results have shown that for plenty of the parameter cases only slight overlap is seen, meaning it would require absolute uncooperative pedestrians to turn the situation into a hazard. This behavior is very unlikely with the exception of children and people with disabilities.
Furthermore, we assumed that pedestrians are generally hidden behind vehicles. While the recent market trend toward SUVs intensifies this problem, it does not hold for many cases. Since the judgment about safety can depend on small distances, seeing a pedestrian when it is next to the bonnet of a car will solve plenty of those situations. An additional "memory" of the map will also help to identify if a location could have become populated while being occluded. Also support by the vehicle-to-X communications could help to close blind spots. A slight alteration of the path could then already give enough room to guarantee freedom of collision.
Case C in particular has shown that there is a clear necessity for cooperation between the traffic participants or an interpretation of movement for singular entities must be done. Still, it needs to be assumed that the pedestrian will not suddenly change direction and move towards the vehicle. Otherwise, it becomes impossible to pass the single pedestrian.
In essence, it can be said that a lightweight, conservative approach as presented is suitable to cope with various situations. Only under the assumption of uncooperative behavior, more sophisticated solutions are necessary. Since these situations are comparable rare, a dynamic selection of methods to assess safety under the viewpoint of maintaining performance is possible. Hence, the need for overly powerful hardware can be eased and the validation task can be narrowed to special cases, drastically reducing the efforts to prove system safety.

\section{Conclusion}
We have shown a combined approach of a variable speed cellular automaton based safety monitor and a nonlinear model predictive controller for safe automated driving in unstructured and partially occluded environments.
The suggested system set-up features deterministic behavior in combination with modest hardware requirements.
By three different experiments we have evaluated the performance of our approach and provided test-data for different parameter combinations.
It has been shown that a conservative prediction is suitable for a wide range of scenarios and environment parameters. Furthermore, we have pointed out the special cases which require more detailed assessment to ensure safety while providing a smooth driving style. This enables to conclude that only a certain set of problems need more sophisticated solutions which can be run on demand, mitigating overly complex hard- and software set-ups.
The given approach offers a perfect basis for such a modular system as the baseline algorithms never compromise on safety but only degrade on performance if no suitable assessment of the situation can be made.

\bibliographystyle{IEEEtran}
\bibliography{bibliography,bibliography2}

\end{document}